\documentclass[12pt]{article}
\usepackage[utf8]{inputenc}
\usepackage{amsmath}
\usepackage{enumitem}
\usepackage{graphicx}
\usepackage{hyperref}
\usepackage[left=2cm, right=2cm, top=2cm, bottom=2cm]{geometry}
\usepackage{amssymb,amsfonts}
\usepackage{algorithmic}
\usepackage{textcomp}
\usepackage{xcolor}
\usepackage[utf8]{inputenc}
\usepackage{algorithm2e}
\usepackage{url}
\usepackage{subcaption}

\title{Robotic Arm Manipulation with Inverse Reinforcement Learning \& TD-MPC }

\author{
    \begin{tabular}{cc}
        Sabir Md Sanaullah & Md Shoyib Hassan \\
        2012499642 & 2011856042 \\
        \textit{Electrical \& Computer Engineering} & \textit{Electrical \& Computer Engineering} \\
        \textit{North South University} & \textit{North South University} \\
        Dhaka, Bangladesh & Dhaka, Bangladesh \\
        \texttt{sabir.sanaullah@northsouth.edu} & \texttt{shoyib.hassan@northsouth.edu}
    \end{tabular}
}

\renewenvironment{abstract}
 {\small\bfseries
  \begin{center}
  \bfseries\abstractname\vspace{-1em}\vspace{0pt}
  \end{center}
  \list{}{%
    \setlength{\leftmargin}{10mm}
    \setlength{\rightmargin}{\leftmargin}%
  }%
  \item\relax}
 {\endlist}

\begin{document}

\maketitle

\begin{abstract}
\noindent
One unresolved issue is how to scale model-based inverse reinforcement learning (IRL) to actual robotic manipulation tasks with unpredictable dynamics.
The ability to learn from both visual and proprioceptive examples, creating algorithms that scale to high-dimensional state-spaces, and mastering strong dynamics models are the main obstacles. In this work, we provide a gradient-based inverse reinforcement learning framework that learns cost functions purely from visual human demonstrations.
The shown behavior and the trajectory is then optimized using TD visual model predictive control(MPC) and the learned cost functions. We test our system using fundamental object manipulation tasks on hardware.
\end{abstract}


\section{Introduction}
Research on learning from demonstrations is booming because it allows robots to quickly acquire new skills. In inverse reinforcement learning (IRL), for example, demonstrations might assist in a number of ways by having the robot attempt to deduce the objectives or reward from the human demonstrator. The majority of IRL techniques call for expensive to obtain demonstrations that link action and state measurements.

With the use of visual examples, we move closer to model-based inverse reinforcement learning for basic object manipulation tasks. It is believed that model-based IRL techniques are more sample-efficient and have the potential to facilitate generalization \cite{Abbeel2004}. However, their model-free equivalents have had greater success so far in robotics applications with unknown dynamics in the actual world \cite{abbeel2004apprenticeship,Boularias2011,Ebert2018}. Model-based IRL still faces the following significant obstacles: An inner and an outer optimization step are the two nested optimization issues that make up model-based inverse reinforcement learning.
Given a cost function and transition model, a policy is optimised by the inner optimisation problem.The majority of earlier research \cite{Finn2016,finn2016guided,Englert2017} presumes that this robot-environment transition model is known; in reality, the robot usually lacks access to such a model. In order for the inner step to optimize a policy that closely aligns with the observed demonstrations, the outer optimization step seeks to maximize the cost function. Measuring the impact of changes in cost function parameters on the resulting policy parameters makes this step very difficult. This optimization step is approximated in previous work [5, 8, 1] by minimizing a manually created distance metric between policy rollouts and demonstrations. Although this approximation makes the outer optimization step feasible, learning the cost function may become unstable as a result.

Our work addresses these issues and makes model-based IRL from visual demos possible. We pre-train a dynamics model so that the robot can anticipate how its actions would alter this low-dimensional feature representation. 1) We train keypoint detectors \cite{Grefenstette2019} that extract low-dimensional vision features from both the robot and human demos. Once
The robot can utilize its own dynamics model to optimize its actions to attain the same (relative) latent-state trajectory after observing a latent-state trajectory from a human demonstration. 2) By differentiating through the inner optimization step, we used an inverse reinforcement learning technique that makes learning cost functions possible. The IRL algorithm is based on the latest developments in gradient-based bi-level optimization \cite{higher}. This technique enables us to calculate the gradients of cost function parameters in relation to the inner loop policy optimization phase, resulting in an optimization process that is more stable and efficient. We assess our method by gathering human examples of fundamental object manipulation tasks, figuring out the cost functions involved, and replicating comparable actions on a Franka Panda.

\section{Literature Review}

In recent years, the study of cost function learning from demonstrations provided by experts has generated significant interest. As a result several approaches have been proposed to tackle the challenge of learning effective cost functions of robot tasks.

\subsection{Foundational Approaches in IRL}

\cite{Abbeel2004} introduced apprenticeship learning via inverse reinforcement learning (IRL), which forms the foundation for many subsequent works. Their proposed method focuses on learning a policy that performs as well as the demonstrator by matching the feature expectations between the expert and the learner. This method has been built upon in many ways. For example, by incorporating deep learning techniques in more complex tasks \cite{Finn2016}.

\subsection{Probabilistic Frameworks}

\cite{Boularias2011} proposed using relative entropy in IRL, that provided a probabilistic framework for the learning process. Such a method helps to regularize the learning process and improves the stability of the learned cost functions. In a similar fashion the KKT approach introduced by \cite{Englert2017} learns cost functions by leverage the Karush-Kuhn-Tucker conditions from optimization theory.

\subsection{Deep Learning Techniques}

\cite{Wulfmeier2017}, provided a significant contribution to this field by applying deep learning techniques to large-scale cost function learning for path planning. Their methodology exemplifies the effectiveness of deep networks at capturing the complex cost structures from high dimensional input spaces.

\subsection{Visual Learning}

\cite{Finn2017} and \cite{Ebert2018} have made noteworthy contributions in the domain of visual learning by developing methods, which allow for the learning from visual data in order to predict future frames and so plan robot motion in accordance. These techniques use convolutional neural networks for processing visual information and then learning predictive models that are used for control.

\subsection{Meta-Learning Algorithms}

Another recent Advancement includes works on meta learning algorithms which aim to improve the generalization and efficiency of learning algorithms. A generalized inner loop meta-learning algorithm by \cite{Grefenstette2019} was introduced that adapts quickly to new tasks by learning and initialization which can be fine-tuned with minimal data.

\subsection{Visual Model Predictive Control}

The capability of optimizing action sequences in order to minimize task cost of a given visual dynamics model is fundamental to many robot applications. Many approaches have been developed to either learn a transition model directly in pixel space or learn it jointly in a latent space encoding with a dynamics model in that space. For instance, methods like those proposed by \cite{finn2016guided} have concentrated on pixel-level transitions models and the design of cost functions in evaluating progress to goal positions and success classifiers. While others have mapped visual observations to a learned pose space and utilized deep dynamics models for optimizing actions \cite{wulfmeier2017large}.

\subsection{Inverse Reinforcement Learning}

It's been proven challenging to scale IRL to real-world manipulation tasks. Prior research has explored model-free approaches, which have shown success in certain applications. \cite{kalakrishnan2013learning} used proprioceptive state measurements without considering visual feature spaces, limiting their applicability to tasks requiring visual inputs. Newer advances introduce things like gradient-based bi-level optimization, allowing for the computation of cost functions gradients as a function of the inner loop policy optimization step, which leads to more stable and effective optimization. \cite{osa2018algorithmic}.

\subsection{Gradient-Based Visual Model Predictive Control Framework}

A new approach involving the training of keypoint detectiors to extract low-dimensional vision features from expert demonstrations and pre-train a dynamcis model allows the robot to predict action outcomes in that feature space. This helps the robot to optimize its actions to achieve the same latent-state trajectory observed in the demonstrations. By differentiating through the inner optimization step, this gradient-based IRL algorithm offers significant improvements over traditional feature-matching IRL methods \cite{finn2017deep}. By differentiating through the inner optimization step, this gradient-based IRL algorithm offers significant improvements over traditional feature-matching IRL methods \cite{abbeel2004apprenticeship}.

\subsection{Applications and Experimental Validation}

The experimental validations on robotic platforms, such as the Kuka iiwa, signify their effectiveness in real-world tasks. The combination of self-supervised data collection and expert controller data significantly enhances the training of keypoint detectors and dynamics models, which leads to a more accurate and reliable cost function learning \cite{wulfmeier2017large}.\\

Cost function learning from demonstrations involves a wide range of techniques, from the traditional IRL methods to advanced deep learning and meta-learning approaches. They holistically contribute to the development of more autonomous and capable robotic systems.

\section{Temporal-Difference Visual Model Predictive Control Framework}

In this section, we describe our temporal-difference visual model predictive control approach that combines recent advances in unsupervised keypoint representations and model-based planning. The IRL framework, with a simplified illustration in Figure 1, actually consists of the following components: 1) a keypoint detector that produces low-dimensional visual representations (Human demos), in the form of keypoints, from RGB image inputs; 2) a dynamics model that takes the current joint state \( q \), previous joint state \(\dot{q}\), and actions \( u \) to predict the keypoints and joint state at the next time step; and 3) a gradient-based visual model-predictive planner that optimizes actions for a given task using the dynamics model and a cost function. We will elaborate on each of these modules next.

\subsection{Keypoints as Visual Latent State and Dynamics Model}

We employ an autoencoder with a structural bottleneck to detect 2D keypoints that correspond to pixel positions or areas with maximum variability in the input data. The keypoint detector's architecture closely follows the implementation in \cite{keypoint_paper}. For training the keypoint detector, we collect visual data \( D_{\text{key-train}} \) for self-supervised keypoint training (refer to Appendix A.2). After this training phase, the keypoint detector predicts keypoints \( z = g_{\text{key}}(o_{\text{im}}) \) of dimensionality \( K \times 3 \). Here, \( K \) is the number of keypoints, and each keypoint is given by \( z_k = (z_{x,k}, z_{y,k}, z_{m,k}) \), where \( z_{x,k} \) and \( z_{y,k} \) are pixel locations of the \( k \)-th keypoint, and \( z_{m,k} \) is its intensity, which corresponds roughly to the probability that the keypoint exists in the image.

Given a trained keypoint detector, we collect dynamics data \( D_{\text{dyn-train}} \) to train the dynamics model. The dynamics model learns to predict the next keypoints \( \hat{z}_{t+1} \) and the next joint state \( q_{t+1} \) based on the current joint state \( q_t \), previous joint state \( \dot{q}_t \), and the action \( u_t \).

\subsection{Temporal-Difference Learning in Model Predictive Control}

Our framework incorporates temporal-difference learning to refine the predicted trajectories. The IRL loss \( L_{\text{IRL}} \) is defined as the squared distance between the demonstrated trajectory \( \tau_{\text{demo}} \) and the predicted trajectory \( \hat{\tau} \):

\[
L_{\text{IRL}}(\tau_{\text{demo}}, \hat{\tau}) = \sum_{t} (z_{t,\text{demo}} - \hat{z}_{t})^2
\]

The optimization problem for the IRL is expressed as:

\[
\nabla_{y} L_{\text{IRL}}(\tau_{\text{demo}}, \hat{\tau}_{y}) = \nabla_{y} L_{\text{IRL}}(\tau_{\text{demo}}, \hat{\tau}_{y}) \nabla_{y} \hat{\tau}_{y}
\]

\[
= \nabla_{y} L_{\text{IRL}}(\tau_{\text{demo}}, \hat{\tau}_{y}) \nabla_{y} f_{\text{dyn}}(s; u_{\text{opt}})
\]

\[
= \nabla_{y} L_{\text{IRL}}(\tau_{\text{demo}}, \hat{\tau}_{y}) \nabla_{y} f_{\text{dyn}}(s; u_{\text{init}}) \nabla_{u} C_{y}(s_{\text{demo}}, f_{\text{dyn}}(s; u))
\]

This optimization involves tracking gradients through the inner loop and differentiating the optimization trace with respect to outer parameters using a gradient-based optimizer such as \texttt{higher} \cite{higher}. The algorithm extends to multiple time steps by adapting the equations to the predicted trajectory over \( T \) time steps.

\section{Gradient-Based IRL from Visual Demonstrations}
In this section, we employ a gradient-based inverse reinforcement learning (IRL) algorithm to learn the cost functions directly from visual demonstrations. This approach leverages the new advances made in gradient-based bi-level optimization in order to facilitate the stable and efficient learning of the cost function. The main components of this method involves a pre-trained visual dynamics model and a differentiable action optimization process. 

Algorithm~\ref{alg:grad-irl} outlines the steps involved in our gradient-based IRL for a single demonstration. It begins by first initializing the action sequence and rolling out the initial trajectory using the pre-trained dynamics model. Then it optimizes the action sequence through minimizing the cost function and rolls out the optimized trajectory and to update the state.

\begin{algorithm}
\caption{Gradient-Based IRL for 1 Demo}
\label{alg:grad-irl}
\begin{algorithmic}[1]
\STATE Initial $\psi$, pre-trained $f_{dyn}$, learning rates $\eta = 0.001, \alpha = 0.01$
\STATE Initialize $u_{init,t} = 0, \forall t = 1, \ldots, T$
\STATE // Rollout using the initial actions
\STATE $\hat{z}_0 = z_0, \hat{\theta}_0 = \theta_0$
\STATE $\hat{\tau} = \{\hat{z}_0\}$
\FOR{$t \leftarrow 1 : T$}
    \STATE $\hat{s}_{t-1} = [\hat{z}_{t-1}, \hat{\theta}_{t-1}, \hat{\dot{\theta}}_{t-1}]^T$
    \STATE $\hat{z}_t, \hat{\theta}_t, \hat{\dot{\theta}}_t = f_{dyn}(\hat{s}_{t-1}, u_{init,t-1})$
    \STATE $\hat{\tau} \leftarrow \hat{\tau} \cup \hat{z}_t$
\ENDFOR
\STATE // Action optimization
\STATE $u_{opt} \leftarrow u_{init} - \alpha \nabla_u C(\hat{\tau}, z_{goal})$
\STATE // Get planned trajectory by rolling out $u_{opt}$
\STATE $\tilde{z}_0 = z_0, \tilde{\theta}_0 = \theta_0, \tilde{\dot{\theta}}_0 = \dot{\theta}_0$
\FOR{$t \leftarrow 1 : T$}
    \STATE $\tilde{z}_t, \tilde{\theta}_t = f_{dyn}([\tilde{z}_{t-1}, \tilde{\theta}_{t-1}, \tilde{\dot{\theta}}_{t-1}], u_{opt,t-1})$
\ENDFOR
\STATE \textbf{return} $\tilde{z}, \tilde{\theta}$
\end{algorithmic}
\end{algorithm}

\subsection{Learning Cost Functions for Action Optimization}
In the case of any IRL, learning the cost function that can accurately reflect the demonstrated behavior is crucial. The approach below utilizes a gradient-based bi-level optimization in order to differentiate through the inner optimization step, allowing the algorithm to update the cost function parameters effectively.

The inner optimization problem can be formulated as follows:
\begin{equation}
u_{opt} = \arg\min_u C(\tau, z_{goal})
\end{equation}
where $\tau$ is the trajectory generated by the dynamics model $f_{dyn}$ and the action sequence $u$. The outer optimization step aims to minimize the IRL loss with respect to the cost function parameters $\psi$:
\begin{equation}
\min_\psi L_{IRL}(\tau_{demo}, \hat{\tau}_\psi)
\end{equation}
where $\tau_{demo}$ is the demonstration trajectory and $\hat{\tau}_\psi$ is the predicted trajectory using the current cost function parameters.

By applying the chain rule, we decompose the gradient of the IRL loss with respect to the cost function parameters and it is as follows:
\begin{equation}
\nabla_\psi L_{IRL}(\tau_{demo}, \hat{\tau}_{C_\psi}) = \nabla_{\hat{\tau}_\psi} L_{IRL}(\tau_{demo}, \hat{\tau}_{C_\psi}) \cdot \nabla_\psi f_{dyn}(s, u_{init} - \eta \nabla_u C_\psi(s_{demo}, f_{dyn}(s,u)))
\end{equation}
The formulation allows the update of the cost function parameters iteratively, leading to an effective optimization process.

\subsection{Cost functions and IRL Loss for learning from visual demonstrations}

The specification of the IRL loss \( L_{\text{IRL}} \) and the cost function parameterization \( C_{\psi} \) are both necessary for our algorithm to work. The \( L_{\text{IRL}} \) calculates the difference between the latent trajectory that is displayed (\(\tau_{\text{demo}}\)) and the one that is expected (\(\hat{\tau}\)). To keep the \( L_{\text{IRL}} \) as basic as possible, we use the squared distance as follows:

\[
L_{\text{IRL}}(\tau_{\text{demo}}, \hat{\tau}) = \sum_{t} (z_{t, \text{demo}} - \hat{z}_{t})^2
\]

This equation represents the difference between the anticipated and demonstrated keypoints at each time step.

As with \cite{Englert2017}, we contrast three different cost function \( C_{\psi} \) parameterizations:

\[
C_{\psi}(\tau, z_{\text{goal}}) = \sum_{k} \left( \phi_{x, k} \sum_{t} (\hat{z}_{t, k}^{x} - z_{\text{goal}, k}^{x})^2 + \phi_{y, k} \sum_{t} (\hat{z}_{t, k}^{y} - z_{\text{goal}, k}^{y})^2 \right)
\]

where \(\hat{z}_{t, k}^{x}\) and \(\hat{z}_{t, k}^{y}\) denote the \(k\)th predicted keypoint at time-step \(t\) in the \(x\) and \(y\) dimensions, respectively.

\section{Adversarial IRL with TDMPC}
\begin{figure}[h]
    \centering
    \includegraphics[width=\linewidth]{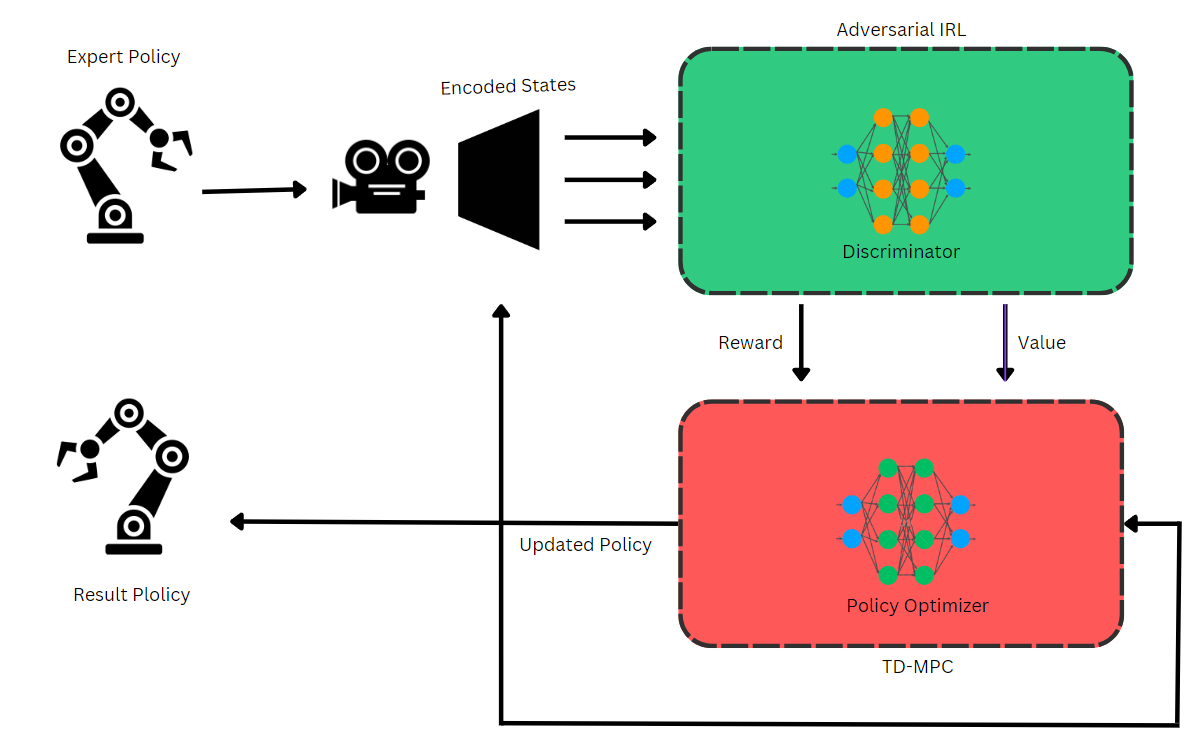}
    \caption{A basic overview of our keypoint-based visual model predictive control framework for AIRL. Actions are optimized via Cross Entropy on the cost function.}
    \label{fig:framework}
\end{figure}

\subsection*{Algorithm: Adversarial Inverse Reinforcement Learning (AIRL)} \subsection*{Initialization} \begin{itemize} \item Initialize the policy network $\pi_\theta$ with parameters $\theta$. \item Initialize the discriminator network $D_\psi$ with parameters $\psi$. \item Initialize the value function $V_\phi$ with parameters $\phi$. \end{itemize} \subsection*{Expert Data} \begin{itemize} \item Collect expert demonstration trajectories $\mathcal{D}_E = \{(s_i, a_i, s'_i)\}$. \end{itemize} \subsection*{Training Loop} \textbf{For each iteration:} \begin{enumerate} \item \textbf{Sample Trajectories} \begin{itemize} \item Sample trajectories $\tau_E$ from expert data $\mathcal{D}_E$. \item Sample trajectories $\tau_\pi$ from the current policy $\pi_\theta$. \end{itemize} \item \textbf{Discriminator Update} \begin{itemize} \item Update the discriminator $D_\psi$ by minimizing the binary cross-entropy loss: \[ \mathcal{L}_D = -\mathbb{E}_{(s, a) \sim \tau_E}[\log D_\psi(s, a)] - \mathbb{E}_{(s, a) \sim \tau_\pi}[\log (1 - D_\psi(s, a))] \] \item Perform gradient descent on $\psi$: \[ \psi \leftarrow \psi - \eta_D \nabla_\psi \mathcal{L}_D \] \end{itemize} \item \textbf{Reward Function} \begin{itemize} \item Define the reward function based on the discriminator output: \[ r_\psi(s, a) = \log D_\psi(s, a) - \log (1 - D_\psi(s, a)) \] \end{itemize}
\item \textbf{Value Function Update} \begin{itemize} \item Update the value function $V_\phi$ using temporal difference learning: \[ \delta_t = r_\psi(s_t, a_t) + \gamma V_\phi(s_{t+1}) - V_\phi(s_t) \] \[ \phi \leftarrow \phi + \eta_V \delta_t \nabla_\phi V_\phi(s_t) \] \end{itemize} 

\item \textbf{Policy Update} \begin{itemize} \item Using the reward function \& value function $r_\psi$ to generate new trajectories, update the policy $\pi_\theta$ using a hybrid reinforcement learning algorithm TD-MPC. 
\item  using TD-MPC: \[ \mathcal{L}_{\pi} = \mathbb{E}_{(s, a) \sim \pi_\theta} [r_\psi(s, a)] \] \[ \theta \leftarrow \theta - \eta_\pi \nabla_\theta \mathcal{L}_{\pi} \] \end{itemize} \end{enumerate} \subsection*{Convergence Check} \begin{itemize} \item Check for convergence criteria (e.g., policy performance threshold, maximum iterations). \end{itemize} \subsection*{Output} \begin{itemize} \item The trained policy $\pi_\theta$ and the inferred reward function $r_\psi$. \end{itemize} \subsection*{Notes} \begin{itemize} \item \textbf{Hyperparameters}: Learning rates $\eta_D$, $\eta_\pi$, and $\eta_V$, discount factor $\gamma$, and batch sizes should be tuned for optimal performance.  \end{itemize}.

\section{Experimental Evaluation}

\subsection{Setup}
We conducted our experiments using a simulated Franka Emika Panda arm in the PyBullet environment. The task involves picking up a small cube, which is spawned at a random position on the ground, and placing it on a larger stationary cube. This setup tests the arm's ability to accurately locate, grasp, and place objects using the learned policy.

\subsection{Data Collection}
We used a collected set of 20 demonstrations by an expert. who manually controlled the Franka Panda arm to complete the task. The demonstrations were recorded at 30Hz and were downsampled to 5Hz. This was done to match the temporal resolution used in the training phase. The keypoints of the arm and the cubes were utilized from each frame to create the dataset.

\subsection{Training}
We trained the model using the dataset to learn the cost function that optimizes the action policy the best, for the placement task. Gradient-based methods were used for 5000 iterations to update the model parameters. 

\subsection{Evaluation}
The performance of the trained model was tested on 10 different scenarios where the small cube was placed at random locations on the floor. The evaluation metrics are the loss and reward with respect to episodes during the training of the model.

\subsubsection{Quantitative Results}
Figure 2(l) shows the performance of the model's loss and reward with respect to the number of episodes it trains through. As can be seen the loss gradually goes down and stabilizes after a certain point and the reward threshold gradually keeps increasing as the training progresses.

\subsubsection{Qualitative Results}
Figure 2(a) to (k) illustrates the placement task at different timesteps using the learned cost functions. As can be seen the robotic franka panda arm is able to perform the task with well suited generalizability and accuracy. Thus, it can be concluded that the model is quite decently effective at performing the given task based on the expert's demonstrations.

\begin{figure}[h]
    \centering
    \begin{tabular}{ccc}
        \includegraphics[width=0.3\linewidth]{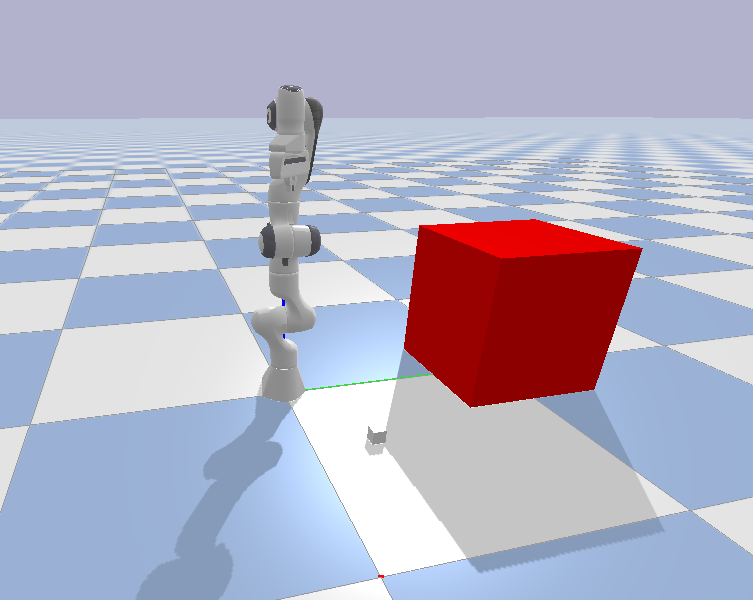} & 
        \includegraphics[width=0.3\linewidth]{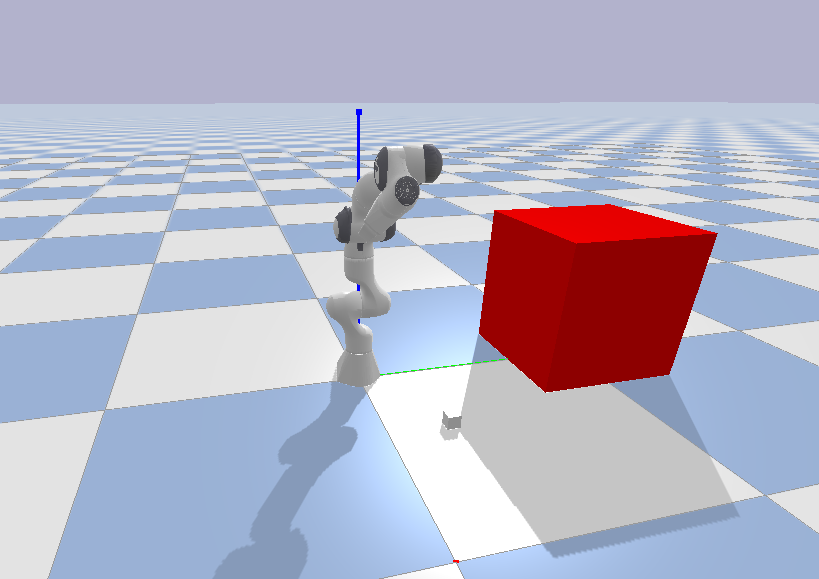} & 
        \includegraphics[width=0.3\linewidth]{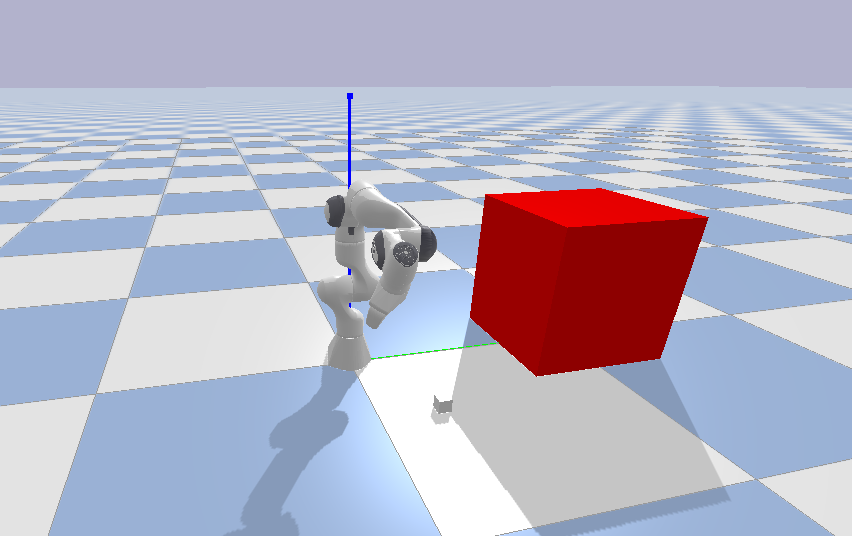} \\
        (a) & (b) & (c) \\
        \includegraphics[width=0.3\linewidth]{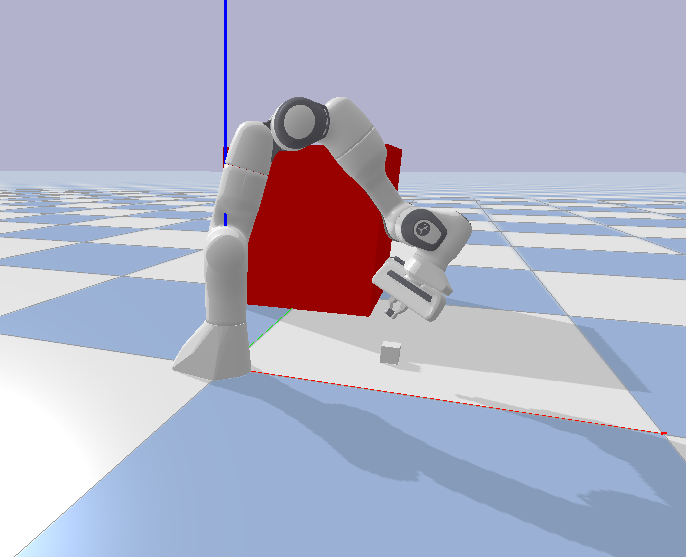} & 
        \includegraphics[width=0.3\linewidth]{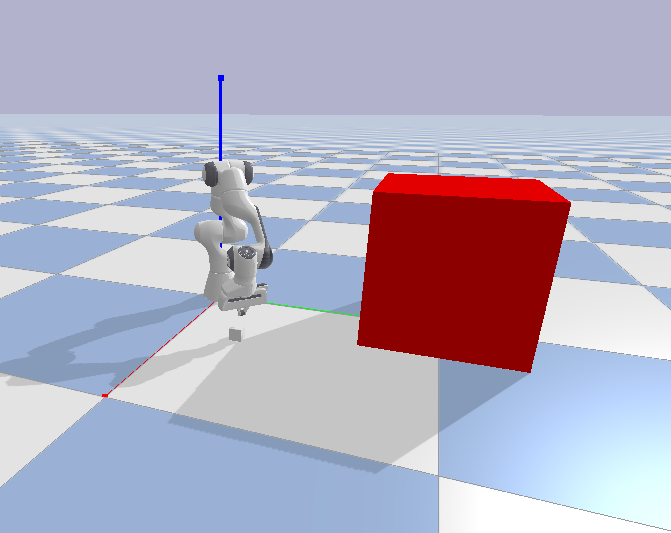} &
        \includegraphics[width=0.3\linewidth]{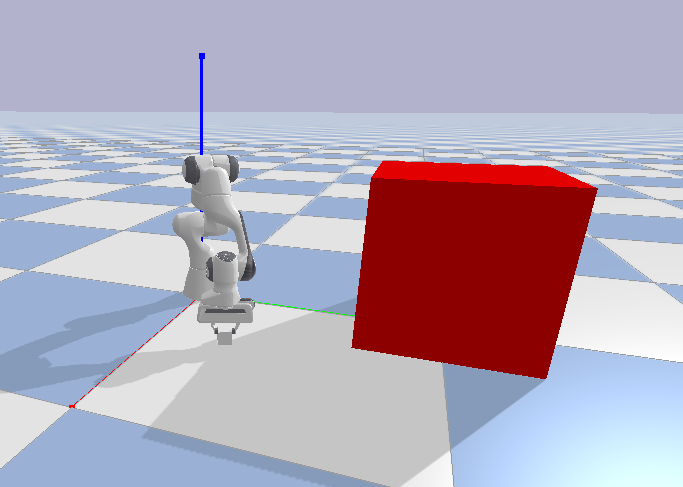} \\
        (d) & (e) & (f) \\
        \includegraphics[width=0.3\linewidth]{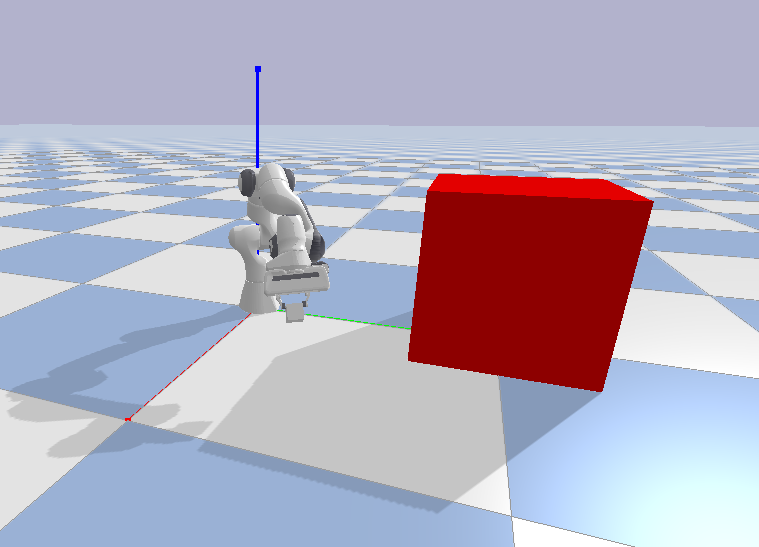} & 
        \includegraphics[width=0.3\linewidth]{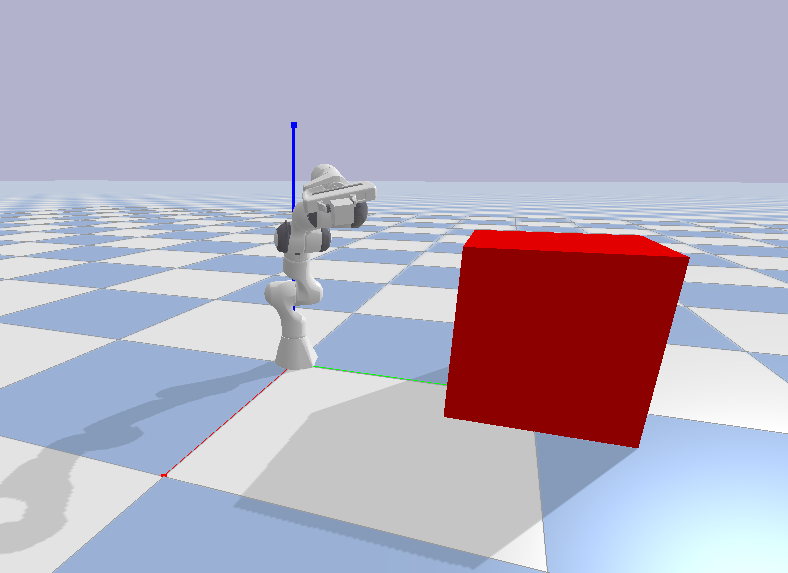} &
        \includegraphics[width=0.3\linewidth]{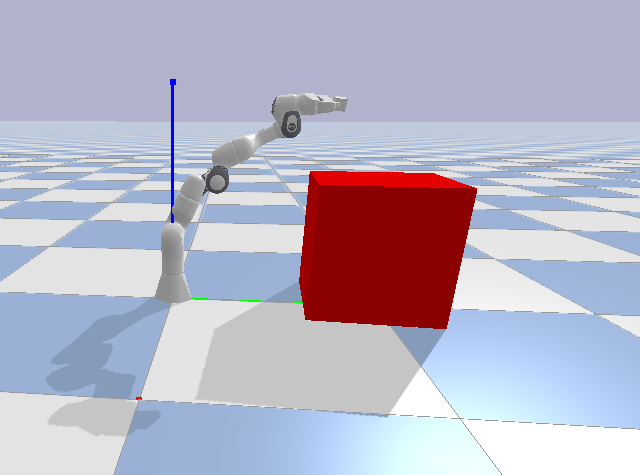} \\
        (g) & (h) & (i) \\
        \includegraphics[width=0.3\linewidth]{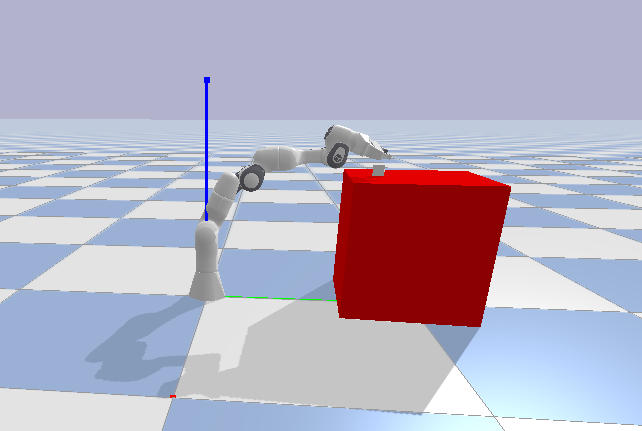} & 
        \includegraphics[width=0.3\linewidth]{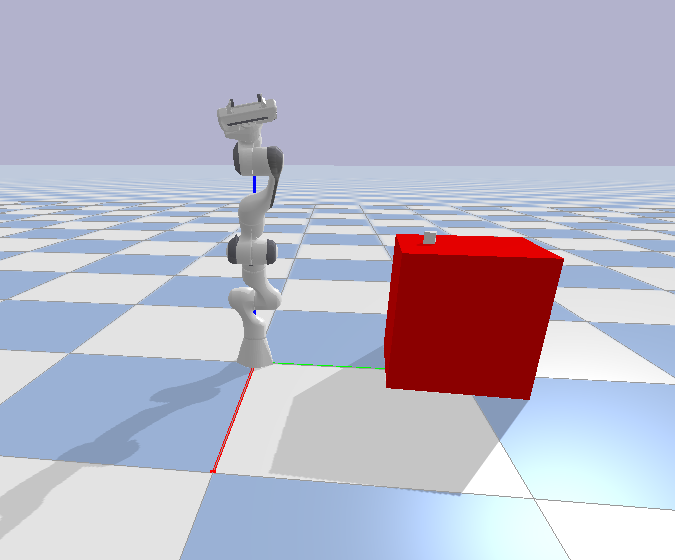} &
        \includegraphics[width=0.3\linewidth]{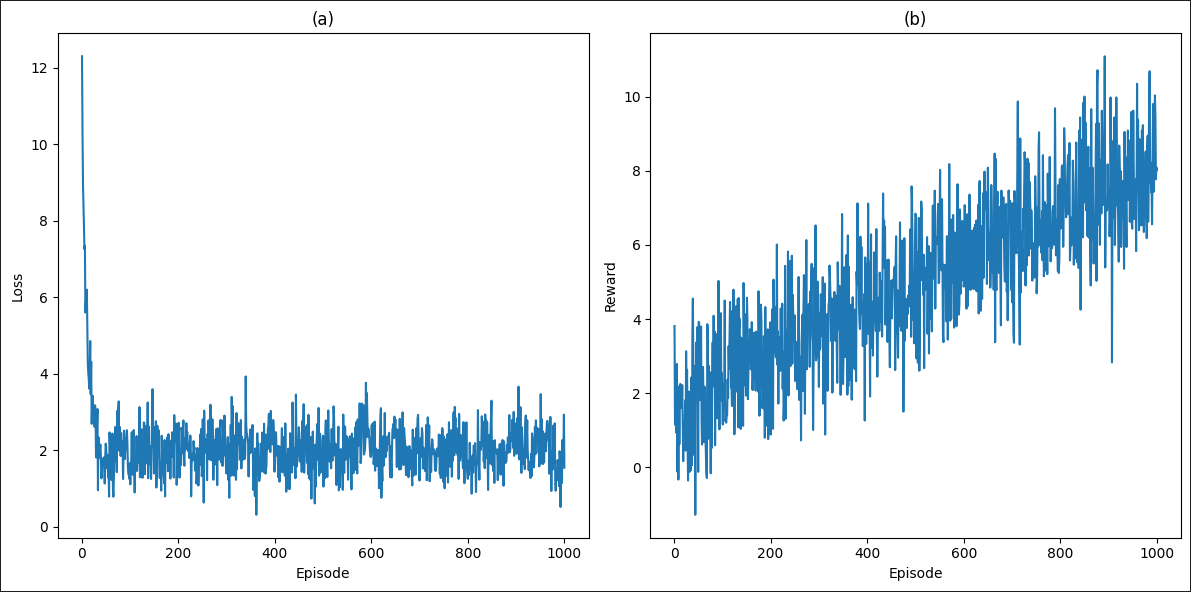} \\
        (j) & (k) & (l) Training statistics \\
    \end{tabular}
    \caption{Figures (a) to (k) represent the progressive performance of the robot in an instance of testing, after training based on the proposed IRL method. Figure (l) represents the change of loss and reward during the training of the model with respect to the number of episodes.}
    \label{fig:results}
\end{figure}

\section{Conclusion and Future Work}
In this paper we proposed a gradient-based IRL framework which learns cost functions from visual demonstrations. Our methodology utilized a compact keypoint-based image representation and trains the visual dynamics model in the latent space. The extracted keypoint trajectories from both the user demos and our learned dynamics model, we've successfully been able to learn different cost functions using the proposed gradient-based IRL algorithm.

The experiment still faces a few challenges. Learning a good visual predictive model is difficult and was a major challenge in this work. One workaround could be to robustify the keypoint detector using methods such as the Florence et al. one, rendering it invariant to different points of view. Moreover, the current approach assumes that demonstrations are from the robot's perspective. And so we addressed the different starting configuarations by learning on relative demos instead of absolute ones. More methods need to be explored in the future so that demonstrations can be better mapped from one context to another, like the case with Liu et al.

And finally, though our experiments show improved convergence behavior for the gradient-based IRL algorithm compared to feature-matching baselines, further investigation is required. An exciting direction for future work is the incorporation of neural network processing (NLP) instructions. By integrating NLP, we could allow users to give commands in natural language which the robot would be able to understand and execute. This incorporation would make the system more user-friendly and more generalizable to a wider range of tasks, enhancing significantly the application of our framework.


\begin{thebibliography}{99}

\bibitem{Abbeel2004}
P. Abbeel and A. Y. Ng, "Apprenticeship learning via inverse reinforcement learning," in \textit{Proceedings of the twenty-first international conference on Machine learning}, 2004, p. 1.

\bibitem{Finn2016}
C. Finn, S. Levine, and P. Abbeel, "Guided cost learning: Deep inverse optimal control via policy optimization," in \textit{International conference on machine learning}, 2016, pp. 49--58.

\bibitem{Boularias2011}
A. Boularias, J. Kober, and J. Peters, "Relative entropy inverse reinforcement learning," in \textit{Proceedings of the Fourteenth International Conference on Artificial Intelligence and Statistics}, 2011, pp. 182--189.

\bibitem{Englert2017}
P. Englert, N. A. Vien, and M. Toussaint, "Inverse KKT: Learning cost functions of manipulation tasks from demonstrations," \textit{The International Journal of Robotics Research}, vol. 36, no. 13-14, pp. 1474--1488, 2017.

\bibitem{Wulfmeier2017}
M. Wulfmeier, D. Rao, D. Z. Wang, P. Ondruska, and I. Posner, "Large-scale cost function learning for path planning using deep inverse reinforcement learning," \textit{The International Journal of Robotics Research}, vol. 36, no. 10, pp. 1073--1087, 2017.

\bibitem{Finn2017}
C. Finn and S. Levine, "Deep visual foresight for planning robot motion," in \textit{2017 IEEE International Conference on Robotics and Automation (ICRA)}, 2017, pp. 2786--2793.

\bibitem{Ebert2018}
F. Ebert, C. Finn, S. Dasari, A. Xie, A. X. Lee, and S. Levine, "Visual foresight: Model-based deep reinforcement learning for vision-based robotic control," \textit{arXiv preprint arXiv:1812.00568}, 2018.

\bibitem{Grefenstette2019}
E. Grefenstette, B. Amos, D. Yarats, P. M. Htut, A. Molchanov, F. Meier, D. Kiela, K. Cho, and S. Chintala, "Generalized inner loop meta-learning," \textit{arXiv preprint arXiv:1910.01727}, 2019.

\bibitem{osa2018algorithmic}
T. Osa, J. Pajarinen, G. Neumann, J. A. Bagnell, P. Abbeel, and J. Peters, "An algorithmic perspective on imitation learning," \textit{arXiv preprint arXiv:1811.06711}, 2018.

\bibitem{kalakrishnan2013learning}
M. Kalakrishnan, P. Pastor, L. Righetti, and S. Schaal, "Learning objective functions for manipulation," in \textit{2013 IEEE International Conference on Robotics and Automation}, 2013, pp. 1331--1336.

\bibitem{finn2016guided}
C. Finn, S. Levine, and P. Abbeel, "Guided cost learning: Deep inverse optimal control via policy optimization," in \textit{International conference on machine learning}, 2016, pp. 49--58.

\bibitem{wulfmeier2017large}
M. Wulfmeier, D. Rao, D. Wang, P. Ondruska, and I. Posner, "Large-scale cost function learning for path planning using deep inverse reinforcement learning," \textit{The International Journal of Robotics Research}, vol. 36, no. 10, pp. 1073--1087, 2017.

\bibitem{abbeel2004apprenticeship}
P. Abbeel and A. Y. Ng, "Apprenticeship learning via inverse reinforcement learning," in \textit{Proceedings of the twenty-first international conference on Machine learning}, 2004, p. 1.

\bibitem{higher}
E. Grefenstette et al., "Higher: A library for differentiable higher-order optimization," in \textit{Neural Information Processing Systems}, 2020.

\bibitem{keypoint_paper}
T. Jakab, A. Gupta, H. Bilen, and A. Vedaldi, "Unsupervised Learning of Object Keypoints for Perception and Control," \textit{Advances in Neural Information Processing Systems}, 2018.

\end{thebibliography}
\end{document}